\documentclass[conference]{IEEEtran}
\IEEEoverridecommandlockouts

\usepackage{float}
\usepackage{cite}
\usepackage{amsmath,amssymb,amsfonts}
\usepackage{algorithmic}
\usepackage{graphicx}
\usepackage{booktabs}
\usepackage{textcomp}
\usepackage{hyperref}
\usepackage{subcaption}
\usepackage{xcolor}
\def\BibTeX{{\rm B\kern-.05em{\sc i\kern-.025em b}\kern-.08em
    T\kern-.1667em\lower.7ex\hbox{E}\kern-.125emX}}
\begin{document}

\bstctlcite{IEEEexample:BSTcontrol}

\title{MANTLE: A Framework for Adaptive In-Situ Planetary Perception Using a Modular Uplink Principle}

\author{\IEEEauthorblockN{Pranav Durai}
\IEEEauthorblockA{\textit{Stanford Center for Innovation in In Vivo Imaging} \\
\textit{Stanford University School of Medicine}\\
318 Campus Drive, Room S040, Stanford, CA 94305, USA \\
}
\and
\IEEEauthorblockN{Gary Doran}
\IEEEauthorblockA{\textit{Jet Propulsion Laboratory} \\
\textit{California Institute of Technology}\\
4800 Oak Grove Drive, Pasadena, CA 91109, USA \\}
}

\maketitle

\begin{abstract}Planetary surface exploration missions rely increasingly on autonomous robotic platforms capable of interpreting complex terrain to ensure safe navigation, enable targeted science, and improve operational efficiency, as demonstrated across past Mars missions from Viking through Perseverance. Among the key perception capabilities, landform classification provides contextual information for landing site selection and scientific analysis, while boulder segmentation supports hazard assessment and path planning. This paper presents MANTLE, a multi-task adaptive network for terrain and landform extraction. The model uses a shared DINOv2 backbone for high-level feature extraction with task-specific heads: a classification head for large-scale landform classification, and a segmentation head for pixel-wise boulder localization, each trained on curated datasets built respectively from HiRISE orbital imagery and MSL surface-level imagery. The classification head achieved a test accuracy of 92.56\% across seven Martian terrain classes, while the segmentation head achieved a validation IoU of 0.753 and showed strong cross-sol generalization on a held-out test set from previously unseen rover traverses. A key advantage of MANTLE is its modular, extensible design, formalized here as the Modular Uplink Principle: only a shared, frozen backbone needs to remain onboard, while subsequent perception capabilities are trained on Earth as lightweight task-specific heads and uplinked without retraining the full model. This work demonstrates two such high-impact capabilities, terrain classification and boulder segmentation, as an initial realization of a framework built to support many more over a mission's lifetime. With this foundation, future explorers need not arrive on Mars fully formed, but can continue to learn, adapt, and grow more capable with every uplink.
\end{abstract}


\section{Introduction}

Robotic exploration remains crucial to understanding the geology, climate, and potential habitability of Mars, as well as other distant planetary bodies in our solar system. Over the past five decades, a succession of surface robotic platforms developed by NASA's Jet Propulsion Laboratory has progressively advanced our ability to explore and interpret the Red Planet. Beginning with the Viking landers in the 1970s, these missions laid the foundation for in-situ analysis of Martian soil and atmosphere. This legacy was further extended through the Mars Pathfinder and its Sojourner rover, which demonstrated the feasibility of semi-autonomous mobility on extraterrestrial terrain. 

Subsequent missions like Spirit, Opportunity, Curiosity, and Perseverance have collectively transformed Mars exploration into a sophisticated exercise in robotic autonomy. Each generation of rover has incorporated increasingly advanced visual perception, navigation, and hazard-avoidance systems, enabling kilometer-scale traverses and detailed geological investigations. 

Perseverance, equipped with the Mastcam-Z~\cite{mastcam} and SuperCam~\cite{supercam} instruments and supported by orbital context imaging from MRO's HiRISE, represents the pinnacle of this evolution, having operated in tandem with the Ingenuity helicopter to scout terrain and plan safe routes. These milestones highlight the growing need for onboard artificial intelligence capable of terrain understanding, scientific target detection, and hazard localization, which are capabilities essential for future surface explorers.

Increasingly, mission concepts require greater onboard autonomy for both in-situ scientific target detection and hazard assessment. To meet these objectives, this work presents MANTLE, a deep learning architecture built around a shared, frozen backbone with lightweight, task-specific heads, capable of operating across both orbital and ground-level vantage points without requiring separate per-modality backbones. In this paper, we show two such heads, one for terrain classification and the other for boulder segmentation, demonstrating that scientific insight and operational safety can be achieved within a single framework.

Unlike separate per-task models, a shared frozen backbone amortizes onboard memory and flight-qualification effort across all current and future perception tasks, so each new capability costs only a lightweight head. The same backbone can in
principle support additional perception tasks through new heads trained independently and added without retraining or modifying it. This marks a step toward fully autonomous, continuously extensible perception for future extraterrestrial orbital and surface missions on Mars and beyond.

The main contributions of this work are:
\begin{enumerate}
    \item A new balanced seven-class high-resolution HiRISE orbital landform dataset \cite{hirise_dataset}, comprising $1024\times1024$ pixel tiles at 0.5~m/pixel.
    \item A new pixel-accurate MSL Mastcam boulder segmentation dataset \cite{msl_dataset} spanning 840 sols, built with zero-shot VLM filtering and semi-automatic SAM\,2~\cite{ravi2025sam} mask generation with human-in-the-loop correction.
    \item MANTLE, a dual-head architecture with a shared frozen DINOv2~\cite{oquab2023dinov2} backbone serving both orbital and ground-level vantage points, comprising a terrain-classification head (92.56\% test accuracy) and a boulder-segmentation head (0.753 validation IoU, 0.735 on a temporally held-out test set).
    \item The Modular Uplink Principle, a formalized deployment model in which new perception capabilities are delivered as KB--MB-scale task-specific head uplinks against a persistent frozen backbone.
    \item Public release of all datasets and training source code\footnote{\url{https://github.com/pranavdurai10/mantle}} to support reproducibility and future extensions.
\end{enumerate}
The remainder of this paper is organized as follows: \autoref{sec:background} reviews related work in Martian terrain classification and hazard detection; \autoref{sec:methodology} describes the two datasets and the MANTLE architecture; \autoref{sec:results} presents experimental results for both heads; \autoref{sec:mup} formalizes the Modular Uplink Principle; and \autoref{sec:conclusion} concludes.

\section{Background}\label{sec:background}

Exploring extraterrestrial terrain comes with many challenges. Automated systems need to process complex visual data to help rovers and landers navigate safely and carry out scientific missions. Machine learning (ML), especially computer vision methods, has become an important tool in solving these problems. By improving techniques like image classification, semantic segmentation, and feature detection, researchers are working to equip autonomous rovers with the ability to study Martian landscapes, find hazards, and identify areas of scientific interest. This work supports NASA's goals of improving robotic exploration and advancing our understanding of other planets.

One fundamental aspect of hazard assessment for mobile robotic platforms is the problem of terrain classification. Ono et al.~\cite{masahiro} explored fully convolutional neural networks~\cite{long} for semantic segmentation of Martian terrain, casting the prediction at each pixel as a multiway softmax classification over a $128\times128$ receptive field and achieving 94.9\% per-pixel classification
accuracy in a single forward pass, though limited annotated ground truth constrained generalization.

Wagstaff et al.~\cite{wagstaff2022impacts} developed an ML-based method to detect fresh impacts on Mars using CTX imagery, achieving 97.5\% accuracy with an Inception-V3 CNN, while related work~\cite{wagstaff_2018_deepmars} applied AlexNet~\cite{krizhevsky_2017_alexnet} to content-based retrieval across the ${\sim}$22-million-image PDS Imaging Atlas. Both works highlight the broader utility of deep learning for science feature
detection in large planetary image archives.

\begin{figure}
    \centering
    \includegraphics[width=1.0\columnwidth]{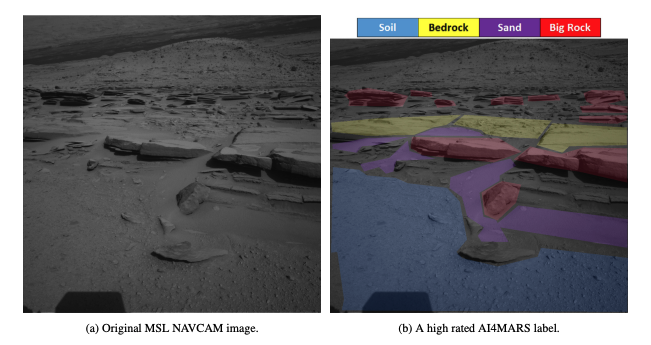}
    \caption{Example annotation from the AI4MARS dataset: (Left) raw MSL Navcam image, and (Right) corresponding polygon-based terrain segmentation labels~\cite{ai4mars}.}
    \label{fig:ai4mars_annotation}
\end{figure}
 
\begin{figure*}[!t]
    \centering
    \begin{subfigure}{0.13\textwidth}
        \includegraphics[width=\linewidth]{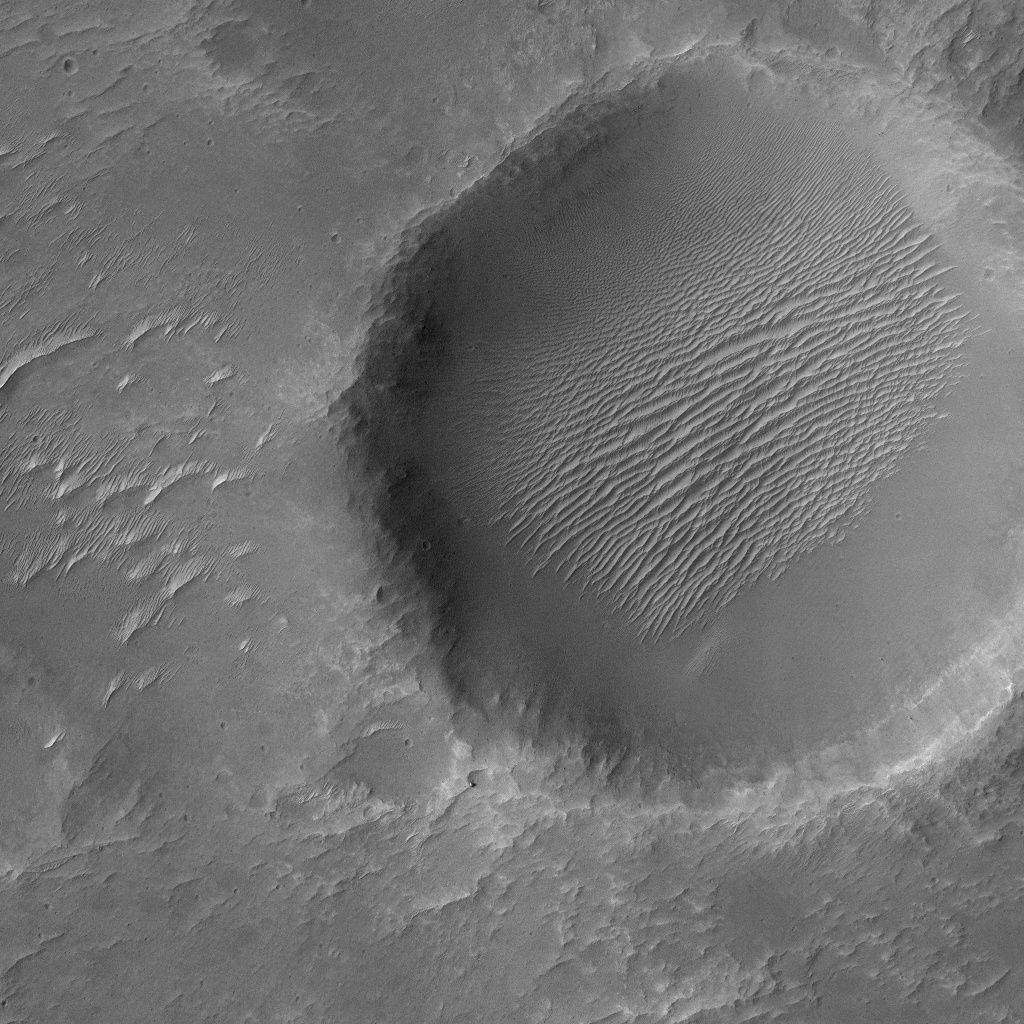}
        \caption{Crater}
    \end{subfigure}
    \hfill
    \begin{subfigure}{0.13\textwidth}
        \includegraphics[width=\linewidth]{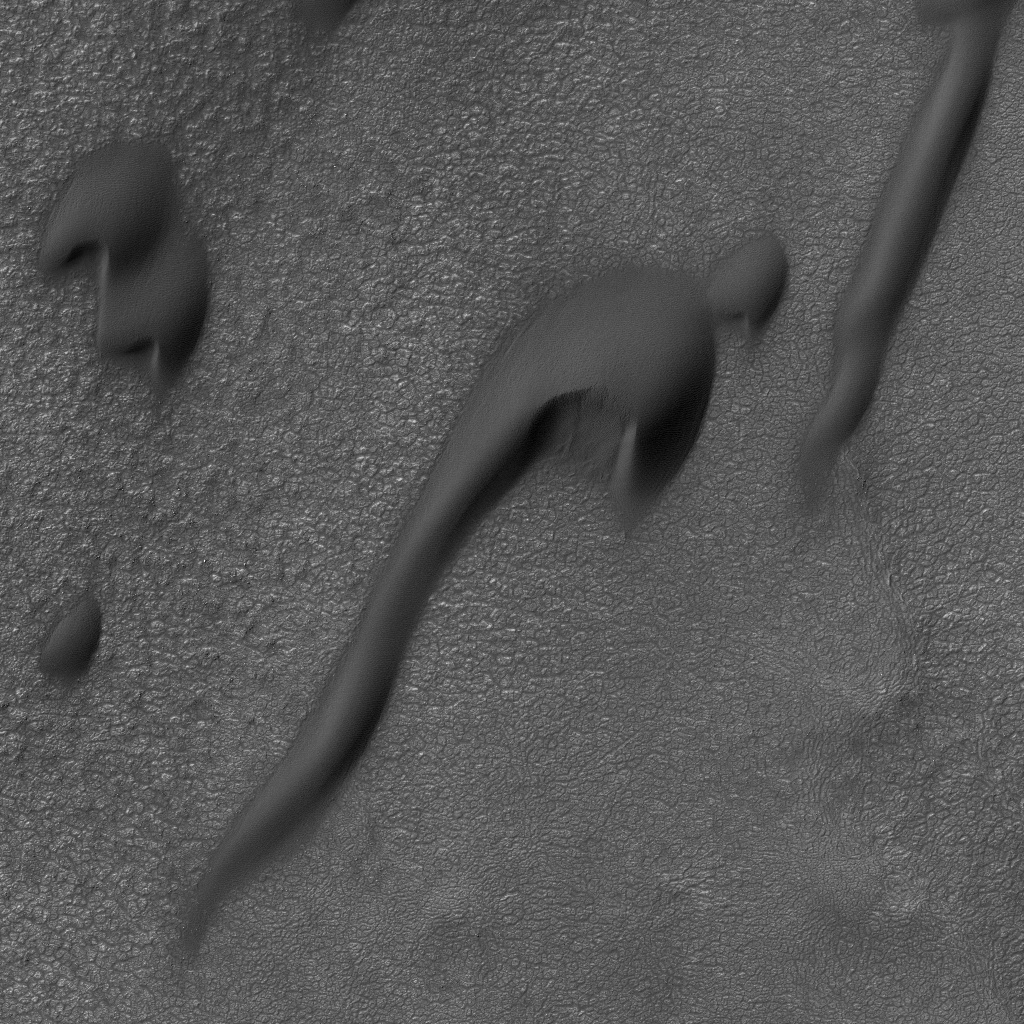}
        \caption{Dark dune}
    \end{subfigure}
    \hfill
    \begin{subfigure}{0.13\textwidth}
        \includegraphics[width=\linewidth]{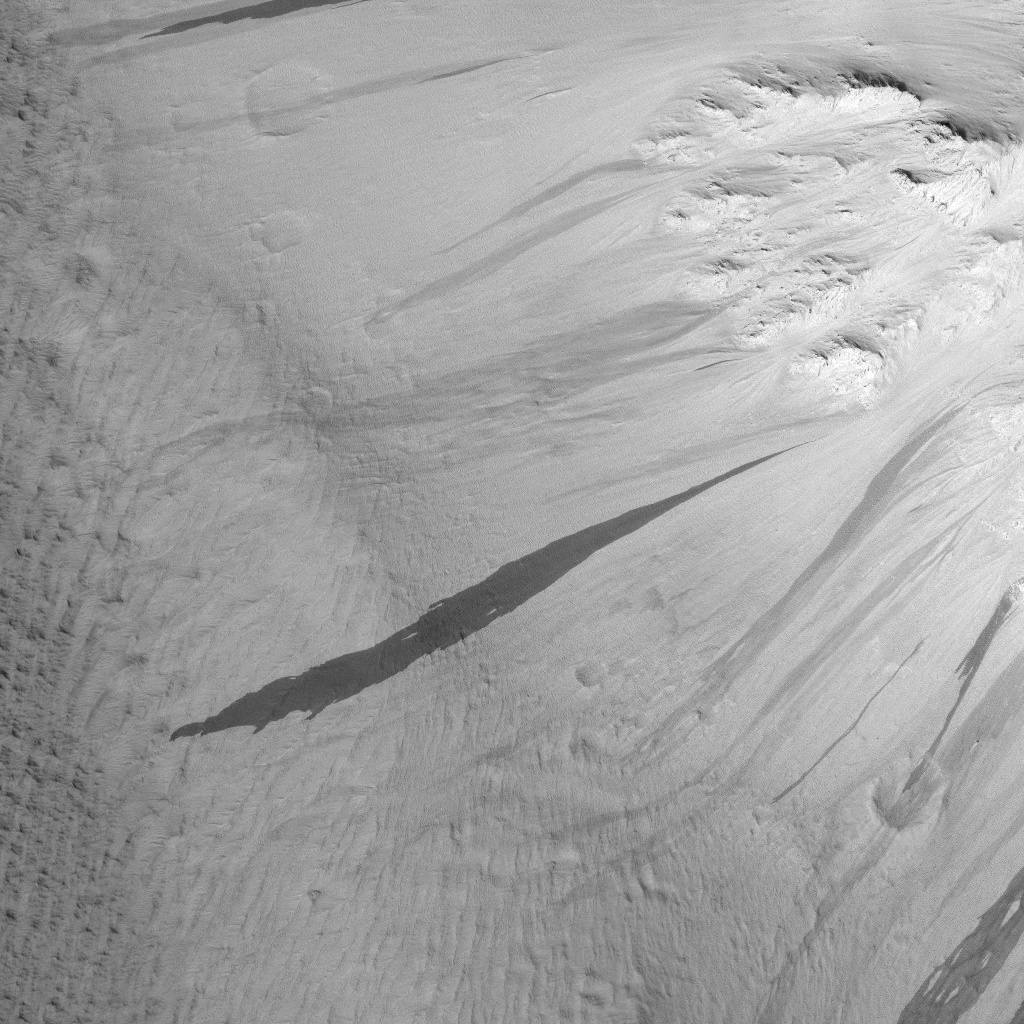}
        \caption{Slope streak}
    \end{subfigure}
    \hfill
    \begin{subfigure}{0.13\textwidth}
        \includegraphics[width=\linewidth]{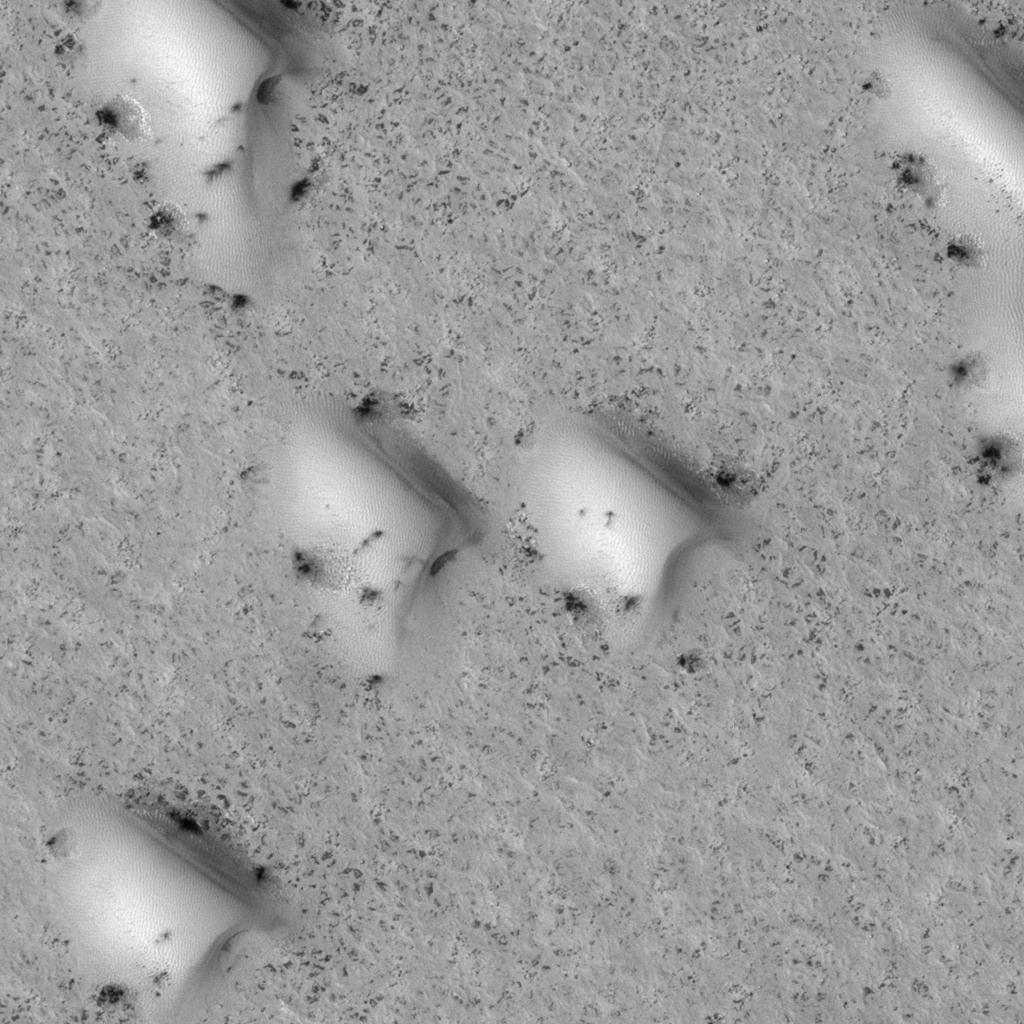}
        \caption{Bright dune}
    \end{subfigure}
    \hfill
    \begin{subfigure}{0.13\textwidth}
        \includegraphics[width=\linewidth]{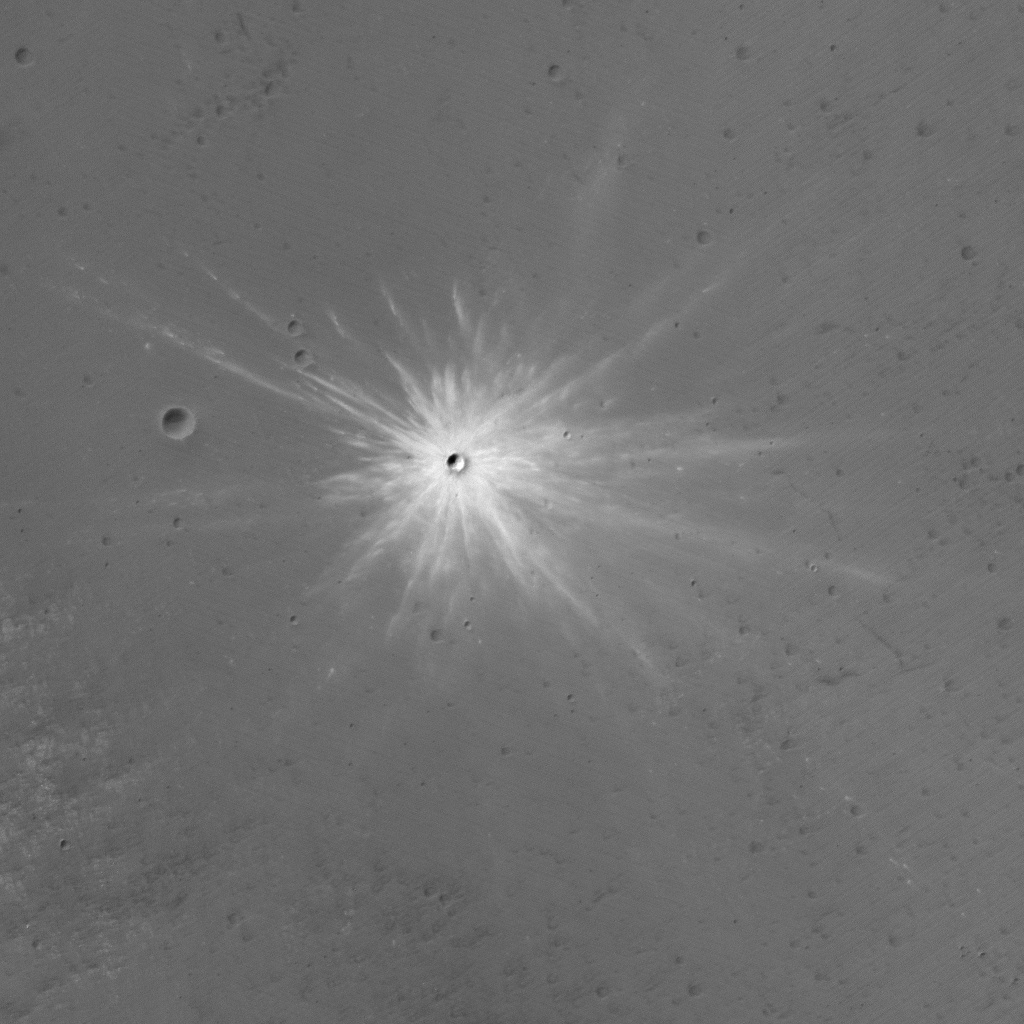}
        \caption{Impact ejecta}
    \end{subfigure}
    \hfill
    \begin{subfigure}{0.13\textwidth}
        \includegraphics[width=\linewidth]{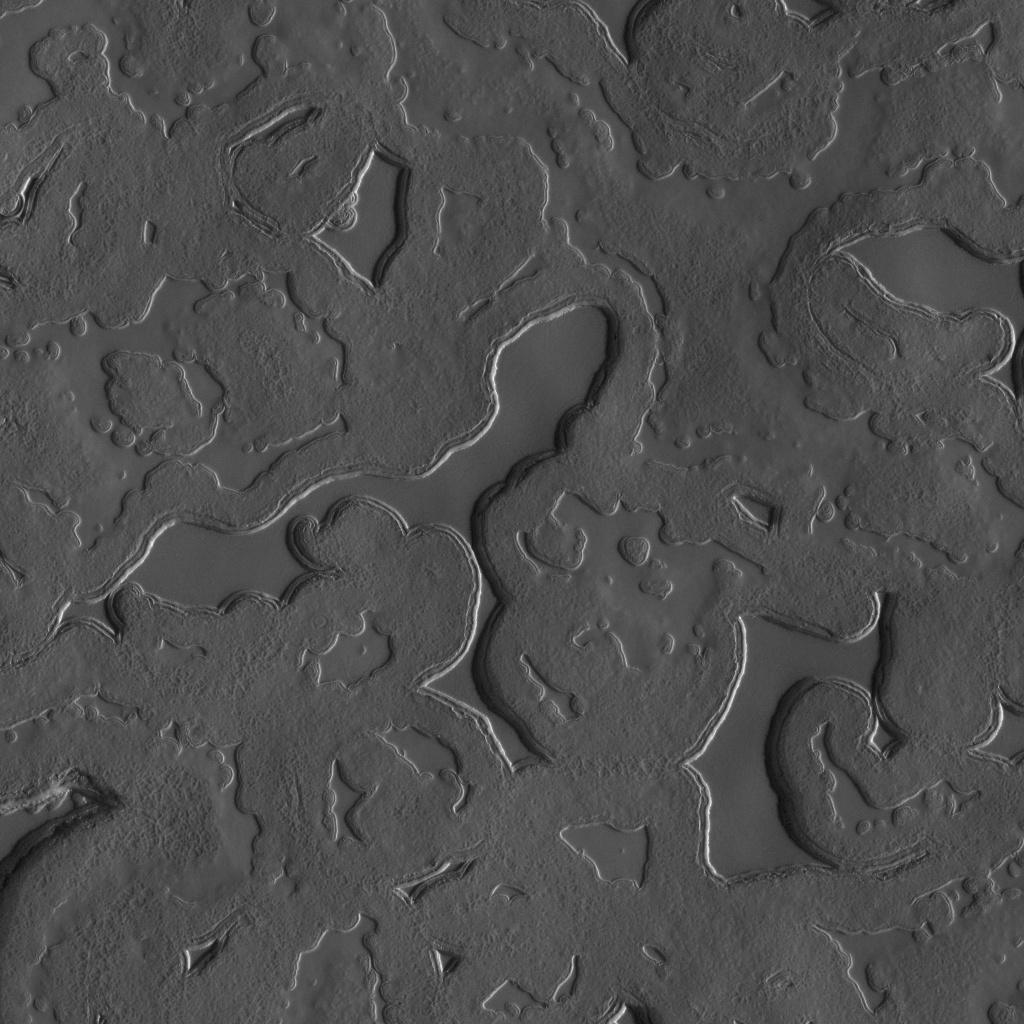}
        \caption{Swiss cheese}
    \end{subfigure}
    \hfill
    \begin{subfigure}{0.13\textwidth}
        \includegraphics[width=\linewidth]{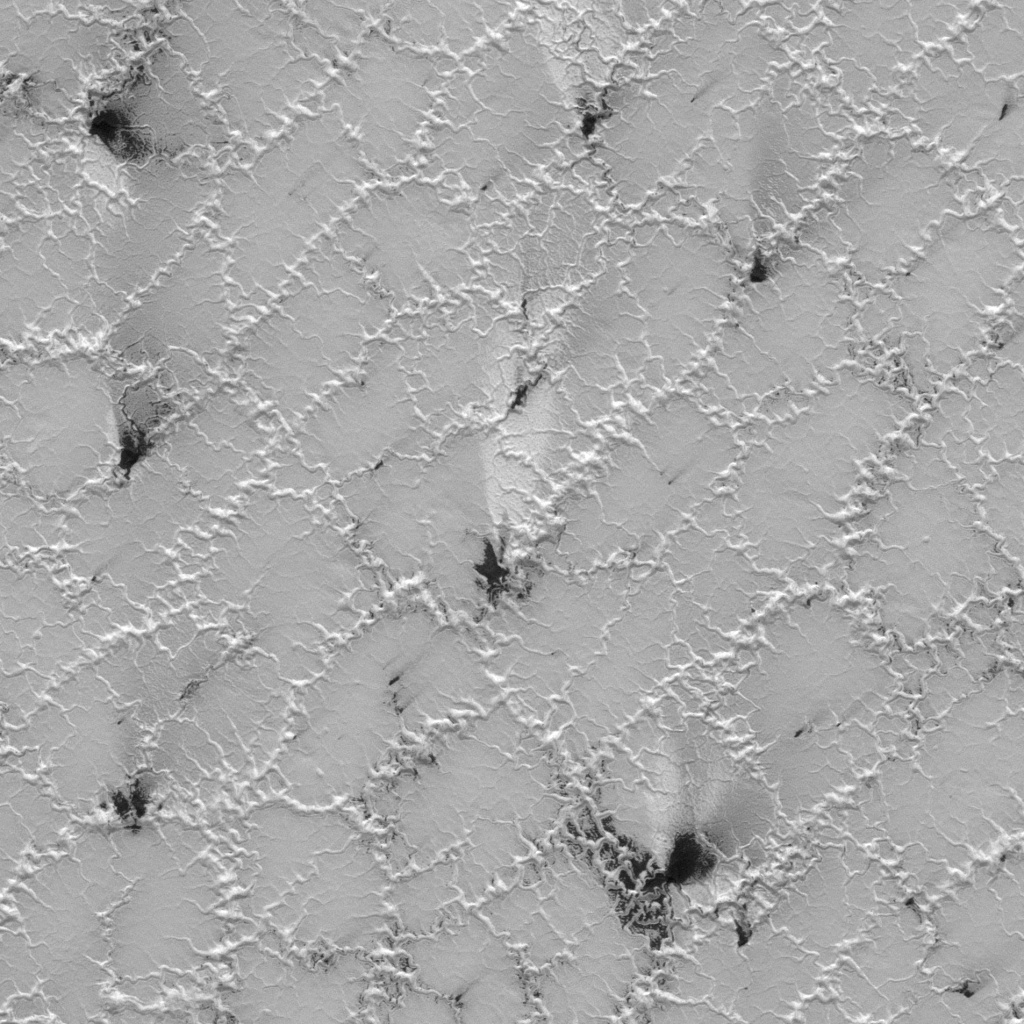}
        \caption{Spider}
    \end{subfigure}
    \caption{Representative HiRISE image patches for each of the seven Martian terrain classes in the High-Resolution HiRISE Landform Dataset.}
    \label{fig:hirise_class_examples}
\end{figure*}

\begin{figure}
    \centering
    \includegraphics[width=1.0\columnwidth]{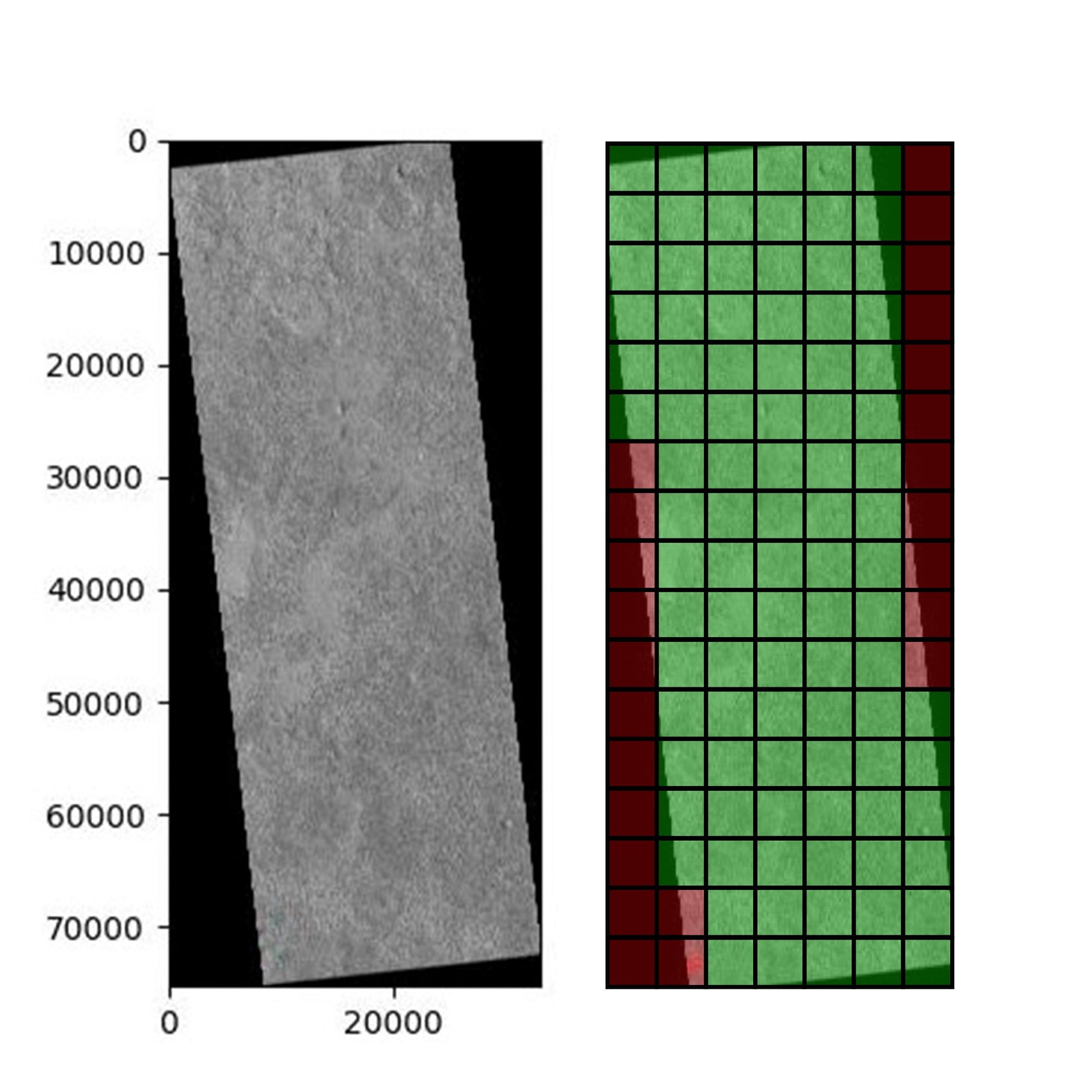}
    \caption{(Left) High-resolution HiRISE image captured by NASA MRO's HiRISE payload: ESP\_011283\_2265\_RED. (Right) Workflow for extracting ROI tiles from HiRISE images, showing the removal of black-pixel-dominated areas and the creation of high-resolution
    $1024\times1024$ pixel patches.}
    \label{fig:hirise_patch_generation_comp}
\end{figure}

\begin{figure*}[t]
    \centering
    \begin{subfigure}{0.23\textwidth}
        \includegraphics[width=\linewidth]{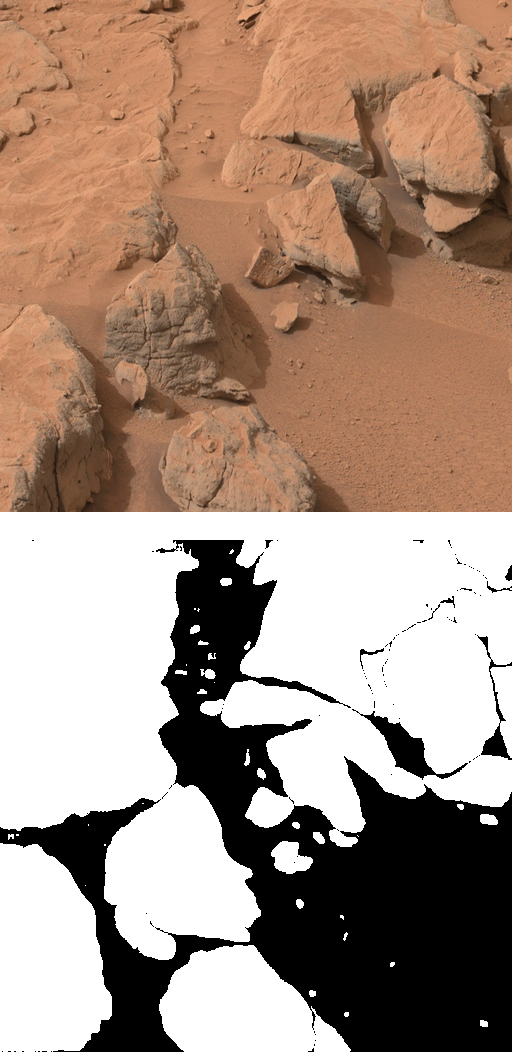}
        \caption{Sol 0132}
    \end{subfigure}
    \hspace{0.002\textwidth}
    \begin{subfigure}{0.23\textwidth}
        \includegraphics[width=\linewidth]{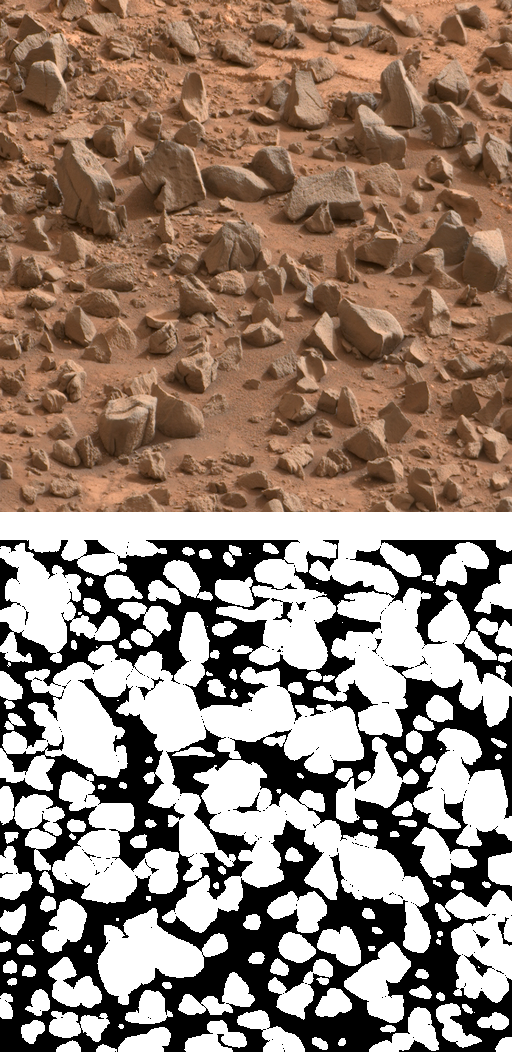}
        \caption{Sol 0844}
    \end{subfigure}
    \hspace{0.002\textwidth}
    \begin{subfigure}{0.23\textwidth}
        \includegraphics[width=\linewidth]{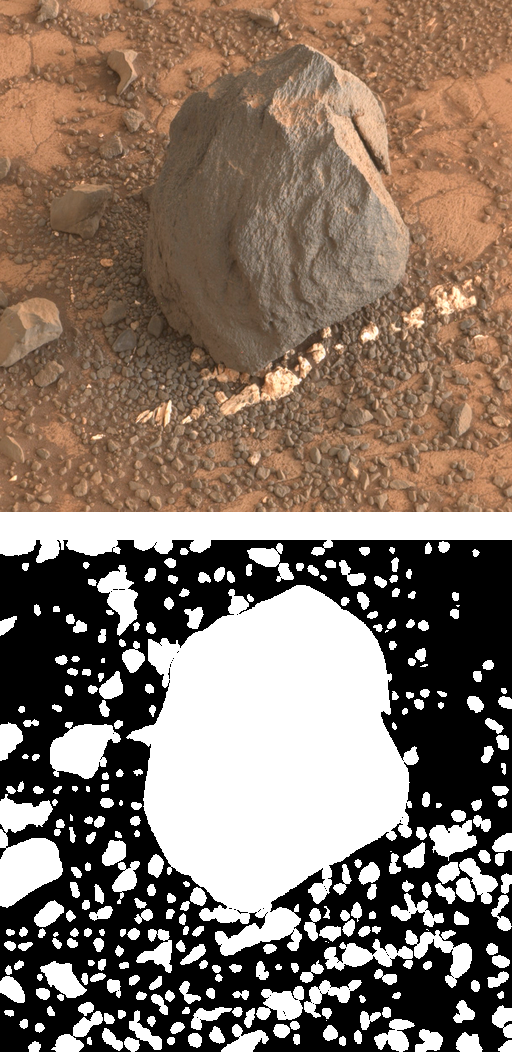}
        \caption{Sol 0925}
    \end{subfigure}
    \hspace{0.002\textwidth}
    \begin{subfigure}{0.23\textwidth}
        \includegraphics[width=\linewidth]{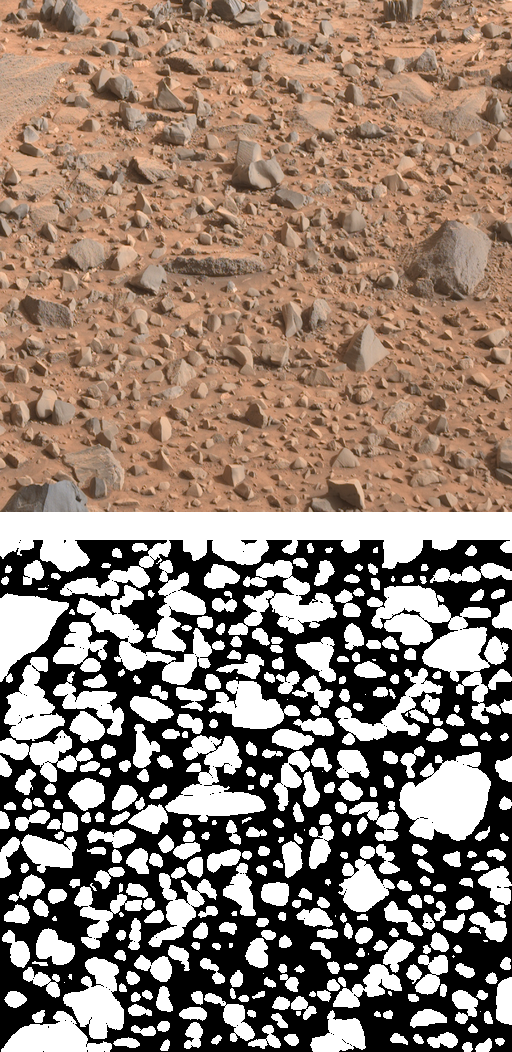}
        \caption{Sol 1081}
    \end{subfigure}
    \caption{Representative image--mask pairs from the MSL Boulder Dataset across different sols, illustrating the diversity of boulder sizes, densities, and terrain contexts captured across the dataset.}
    \label{fig:msl_boulder_dataset_examples}
\end{figure*}

\begin{figure*}[t]
    \centering
    \includegraphics[width=1.0\textwidth]{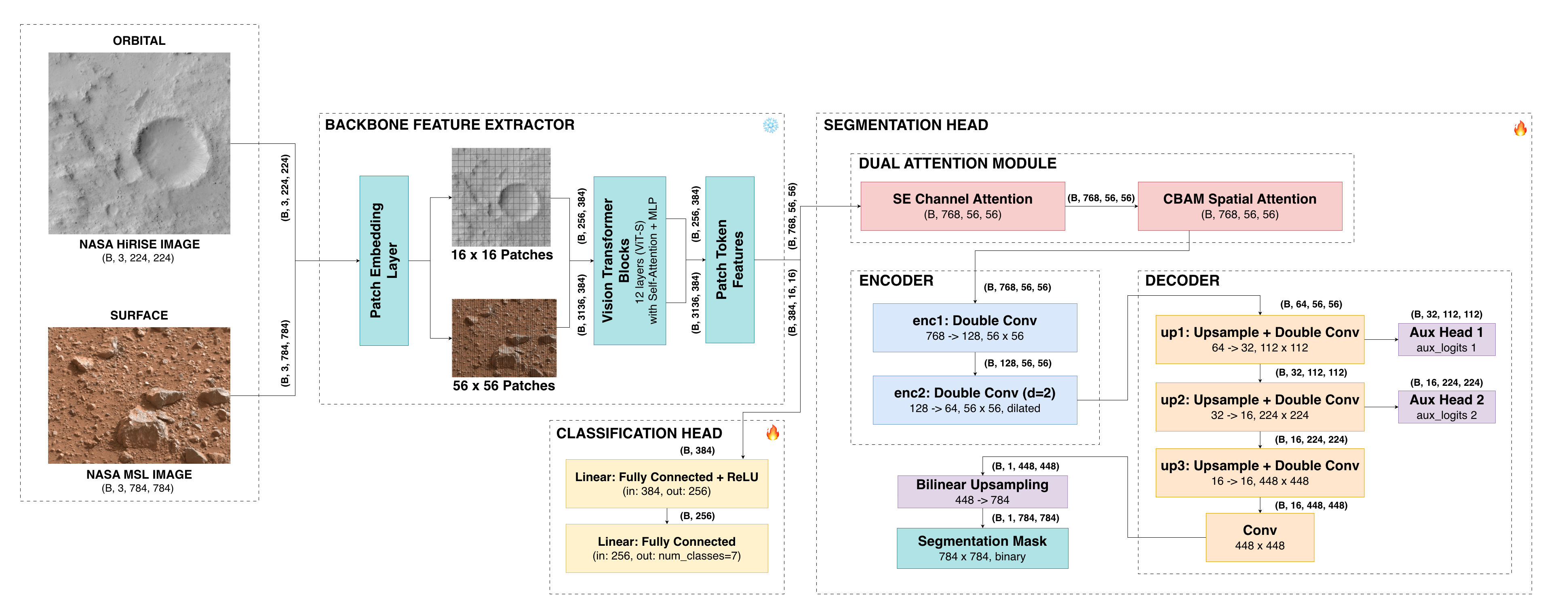}
    \caption{Proposed MANTLE architecture: a shared frozen DINOv2 ViT-S/14 backbone (snowflake) feeds two independently trained task-specific heads (flames denote trainable modules). Orbital HiRISE tiles are resized to $224\times224$ ($16\times16$ patch tokens) for the terrain classification head, while ground-level MSL images are resized to $784\times784$ ($56\times56$ patch tokens) for the boulder segmentation head. Each head is trained separately against the frozen backbone, so new heads can be added without retraining or modifying the backbone.}
    \label{fig:mantle_whole_architecture}
\end{figure*}

Other prior work has focused on the detection of specific hazards, such as large boulders and rocks. Golombek et al.~\cite{golombek} studied the size-frequency distribution of rocks on the northern plains of Mars using high-resolution HiRISE imagery to support site selection for the Phoenix mission, employing an automated shadow-segmentation and geometric modeling pipeline to identify and measure over 10 million rocks across a 1500~km\textsuperscript{2} area with sub-meter accuracy. Hood et al.~\cite{hood} introduced the Martian Boulder Automatic Recognition System (MBARS), an open-source toolkit that similarly uses shadow segmentation on HiRISE imagery (${\sim}$25~cm/pixel), modeling boulders as spheroids and fitting ellipses to shadow boundaries via orthogonal distance regression, achieving precision within $\pm$37.5~cm. Both approaches demonstrated that boulder distributions extracted from orbital imagery align well with lander-based ground truth, establishing shadow-based geometric methods as a reliable foundation for orbital hazard assessment, though their dependence on consistent solar illumination geometry limits applicability to ground-level rover imagery. Stoken et al.~\cite{stoken2023automated} extended this line of work to the lunar surface, training a YOLOv5-based~\cite{ultralytics2021yolov5} network on Lunar Reconnaissance Orbiter NAC imagery to detect boulder shadows and estimate heights for Artemis landing site assessment, illustrating the broader shift toward learning-based detection pipelines.

Prieur et al.~\cite{prieur2023boulder} introduced BoulderNet, a Mask RCNN~\cite{maskrcnn}-based instance segmentation model trained on over 30,000 manually digitized boulders across Earth, Moon, and Mars satellite imagery, achieving detection accuracy comparable to human mappers with morphometric errors within $\pm$15\% of field measurements. While BoulderNet demonstrated that learned segmentation generalizes across planetary bodies and illumination conditions, it works exclusively on orbital imagery and does not address ground-level rover perception. Xiong et al.~\cite{10210278} introduced a transformer-based semantic segmentation framework for Martian rock scenes, combining a Mix Transformer encoder with a feature enhancement module and window transformer block to capture multi-scale local and global context on synthetic and real ZhuRong~\cite{zhurong} rover imagery. While demonstrating strong performance on semantic rock segmentation, their approach targets a fixed segmentation capability on a separate dataset and does not address the modular, uplink-compatible multi-task design unique to the present work.

Closer to the ground-level setting addressed in this work, Swan et al.~\cite{ai4mars} introduced AI4MARS, a large-scale terrain segmentation dataset comprising ${\sim}$326K crowdsourced semantic segmentation labels on 35K images from the Curiosity, Opportunity, and Spirit rovers (each image labeled by ten annotators), plus ${\sim}$1.5K validation labels annotated by MSL and MER rover planners and scientists, developed at NASA's Jet Propulsion Laboratory to support autonomous traversal and path planning (see \autoref{fig:ai4mars_annotation}). While the crowdsourced polygon-based annotations made large-scale labeling tractable, they produced imprecise object boundaries, particularly for small, irregularly shaped boulders, since no tool at the time offered pixel-accurate mask generation at that scale. The coarse, imprecise polygon boundaries visible in \autoref{fig:ai4mars_annotation} directly motivate this work's use of SAM\,2~\cite{ravi2025sam} to produce pixel-accurate boulder masks. Building on this, Dai et al.~\cite{segmarsvit} proposed SegMarsViT, a lightweight ViT-based~\cite{dosovitskiy2020image} terrain segmentation network evaluated on AI4MARS~\cite{ai4mars} and MSL-Seg, targeting onboard deployment under the strict computational and power constraints of rover processors. While SegMarsViT demonstrates that efficient onboard segmentation is feasible, it addresses the deployment constraint through model compression rather than the modular, uplink-extensible architecture proposed here. The present work advances both directions: utilizing SAM\,2~\cite{ravi2025sam} with human-in-the-loop correction to generate pixel-accurate boulder masks at scale, and introducing a shared frozen backbone that supports new perception capabilities through lightweight uplinked heads rather than requiring full model retraining or compression.

Collectively, this body of work reflects two largely separate trajectories in Mars perception research: orbital-scale hazard and landform analysis versus ground-level terrain understanding for autonomy, and, to the best of our knowledge, no prior work has unified both within a single shared backbone operating across both vantage points without separate per-modality models. Furthermore, existing systems treat perception capabilities as fixed at deployment, leaving open the question of how a surface robotic platform, such as a rover, might acquire new visual skills after landing under the severe bandwidth constraints of deep-space operations.

\section{Methodology}\label{sec:methodology}
Exploration of the Martian surface presents unique challenges that require innovative solutions to ensure mission success and scientific discovery. Two critical tasks in this domain are identifying scientifically valuable terrain types and detecting hazardous boulders that pose risks during navigation sequences. Addressing these challenges is essential for mission planning, site selection, and rover safety, but achieving high accuracy for both tasks requires advanced methodologies and robust datasets.

To meet these needs, we present MANTLE, a deep learning architecture built around a shared, frozen backbone with lightweight task-specific heads, designed to perform both scientific terrain classification and boulder hazard detection. Building on recent advances in transformer-based perception, MANTLE demonstrates strong performance on both tasks while handling the unique visual characteristics of Martian imagery. The model is trained on two high-resolution datasets specifically curated for these tasks, one from HiRISE orbital imagery and one from MSL ground-level imagery, providing the foundation for robust and reliable performance across both vantage points. This section presents detailed descriptions of the datasets, followed by the MANTLE architecture.

\subsection{High-Resolution HiRISE Landform Dataset}

The terrain classification dataset builds on the HiRISENet dataset originally compiled by Doran et al.~\cite{gary_doran_2020}, an expanded version of the dataset used for training HiRISENet~\cite{you_lu_hirise_2017} containing 73,031 landmarks extracted from 180 HiRISE browse-quality images with augmentation. The original dataset presented several limitations: severe class imbalance (the \textit{other} class alone containing nearly 40,000 samples), high intra-class variability in the \textit{spider}, \textit{slope}, and
\textit{streak} classes, and low spatial resolution ($227\times227$ tiles from browse-resolution images at ${\sim}10$~m/pixel).
 
To overcome these challenges, we constructed a high-resolution dataset at 0.5~m/pixel with $1024\times1024$ pixel tiles extracted from raw HiRISE images available through the University of Arizona's HiRISE platform~\cite{uahirise}. A sliding window tile extraction algorithm was applied to generate balanced, high-quality patches across seven terrain classes: \textit{crater}, \textit{dark dune}, \textit{slope streak}, \textit{bright dune}, \textit{impact ejecta}, \textit{Swiss cheese}, and \textit{spider} (see \autoref{fig:hirise_class_examples}). Black pixel regions inherent to raw HiRISE images were identified and mitigated via a threshold-based replacement strategy, and each patch was resized to $1024\times1024$ pixels using cubic interpolation. Class balance was fixed at $N_c = 350$ samples per class, with augmentation applied to under-represented classes and pruning applied to over-represented ones. 

Full pipeline details and parameters for the extraction, black pixel handling, and balancing procedures are provided in Supplementary Section~S1.

See \autoref{fig:hirise_patch_generation_comp} for an illustration of the patch extraction workflow.

\subsection{High-Resolution MSL Boulder Dataset}

Boulders pose significant challenges for autonomous planetary surface navigation, where accurate detection is critical for hazard avoidance, path planning, and wheel entrapment prevention. Beyond navigation safety, boulder analysis provides key insights into geological processes such as impact cratering, surface erosion, and material transport, supporting both operational and scientific mission objectives.
 
Raw imagery was sourced from the MSL Curiosity rover's Mastcam instrument via the NASA Planetary Data System\footnote{\url{https://pds-imaging.jpl.nasa.gov/}}; the primary corpus, used for training and validation, was sampled across Sol~0017 through Sol~2302 ($N_{\text{sol}} = 840$ unique sols). The raw corpus of $N_{\text{raw}} = 27{,}122$ PNG images (${\approx}48$~GB) was passed through a four-stage pipeline: (1) zero-shot VLM-based filtering using the Qwen3.5-35B-A3B model~\cite{qwen3.5} to retain only boulder-positive frames, (2) spectral and resolution quality filtering, (3) $4{:}3$ aspect ratio normalization with a $5\%$ border crop, and (4) semi-automatic pixel-accurate mask generation via SAM\,2~\cite{ravi2025sam} with human-in-the-loop correction. 

Full pipeline details and parameters for all pipeline stages are provided in Supplementary Section~S1.
 
Following mask generation and quality control, the final training-and-validation dataset contains $N_{\text{final}} = 6{,}232$ image--mask pairs from 840 unique sols, yielding an average of $7.42$ images per sol. This dataset was partitioned using a sol-level randomized split (80/20) to prevent within-sol leakage, with all images from a given sol assigned exclusively to one partition, yielding $4{,}892$ training and $1{,}340$ validation images across Sol~0017--2302. An additional, temporally held-out test set was sampled from Sol~$\geq$~2305 and processed and annotated with the same pipeline, forming a temporal split to evaluate generalization to new terrains. See \autoref{fig:msl_boulder_dataset_examples} for representative image--mask pairs across different sols.

\subsection{The MANTLE Model}

The MANTLE model addresses the two tasks defined above through a single DINOv2-based transformer~\cite{oquab2023dinov2} backbone feeding two specialized heads: a terrain classification head operating on orbital HiRISE imagery to identify areas of scientific interest, and a boulder segmentation head operating on MSL ground-level imagery to support rover sample analysis and autonomous navigation.
The backbone is kept frozen to enable the uplink-based extensibility formalized in~\autoref{sec:mup}: heads can be trained independently on Earth and deployed against the unchanged onboard backbone.

Refer to \autoref{fig:mantle_whole_architecture} for the consolidated architecture.

\subsubsection{Backbone Feature Extractor}

MANTLE accepts two distinct input modalities that are processed independently through the same frozen backbone.

For terrain classification, each $224 \times 224$ HiRISE input is divided into non-overlapping $14 \times 14$ patches, yielding a $16 \times 16$ grid of $256$ patch tokens. These are projected into a $384$-dimensional embedding space, passed through the 12 ViT blocks of the frozen backbone, and reshaped into a $384 \times 16 \times 16$ spatial feature grid. For boulder segmentation, each $784 \times 784$ MSL input is processed identically, yielding a denser $56 \times 56$ grid of $3{,}136$ tokens; after the ViT blocks, the token sequence is reshaped into a spatial grid and projected to double the channel depth via a learned linear projection, producing a $768 \times 56 \times 56$ feature grid.
The higher-resolution input ($784$ vs.\ $224$) thus provides the segmentation head with $12.25\times$ more spatial tokens and finer-grained localization cues for boulder segmentation. The full tensor-level specification of the feature extraction pipeline is provided in Supplementary Section~S2.

See \autoref{fig:shared_backbone_features} for a visualization of the spatial feature representations extracted by the shared backbone across both input modalities.
In the orbital HiRISE example, the crater emerges as a coherent low-magnitude region sharply delineated from the surrounding plains, indicating that the frozen features already isolate the landform that the classification head must recognize. In the surface MSL example, the activation pattern traces the outlines and contact edges of the foreground boulders, separating them from the surrounding regolith at the resolution the segmentation head consumes. That the same frozen weights organize both a nadir orbital scene and an oblique ground-level scene around their task-relevant structures, without any Mars-specific fine-tuning, is the empirical basis for the paper's premise that a single shared backbone can serve both vantage points.

\begin{figure}[t]
    \centering
    \begin{subfigure}{0.47\columnwidth}
        \includegraphics[width=\linewidth]{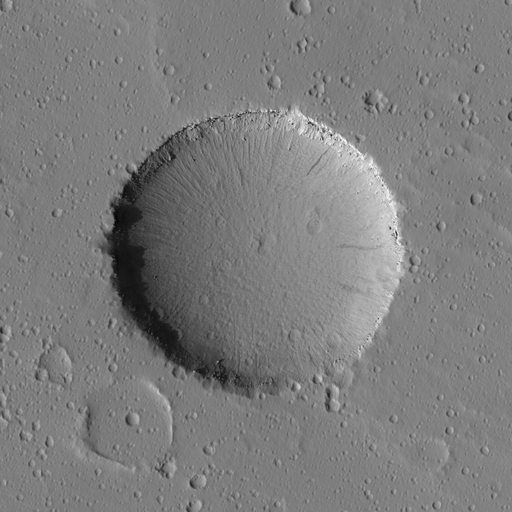}
        \caption{Orbital: HiRISE}
    \end{subfigure}
    \hspace{0.01\columnwidth}
    \begin{subfigure}{0.47\columnwidth}
        \includegraphics[width=\linewidth]{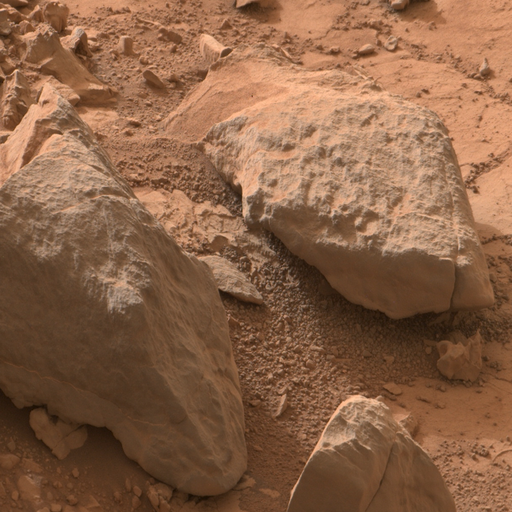}
        \caption{Surface: MSL}
    \end{subfigure}

    \vspace{0.3em}

    \begin{subfigure}{0.47\columnwidth}
        \includegraphics[width=\linewidth]{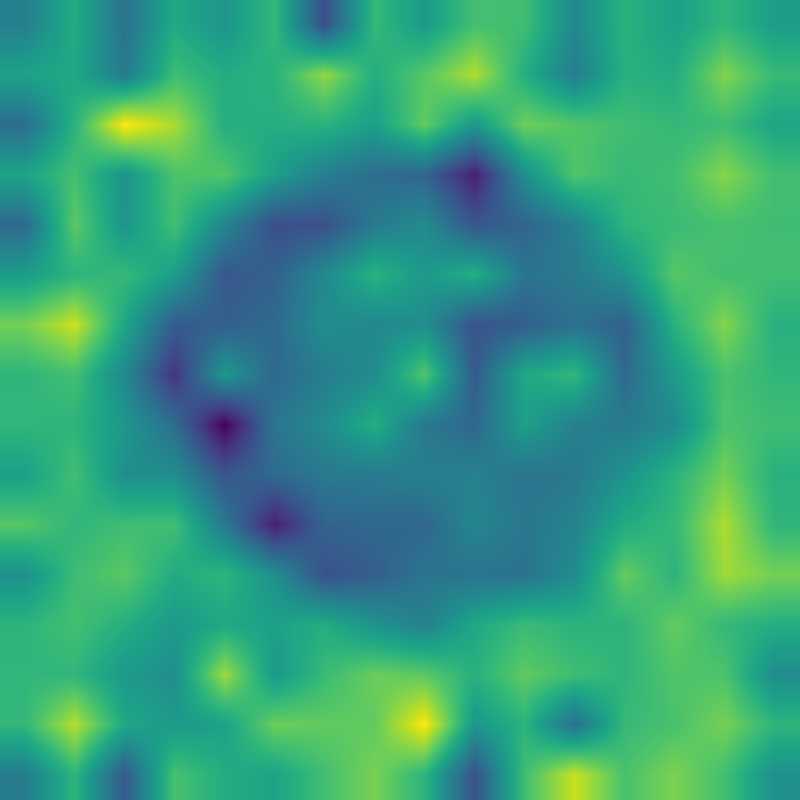}
        \caption{Feature Activation: ($16\times16$)}
    \end{subfigure}
    \hspace{0.01\columnwidth}
    \begin{subfigure}{0.47\columnwidth}
        \includegraphics[width=\linewidth]{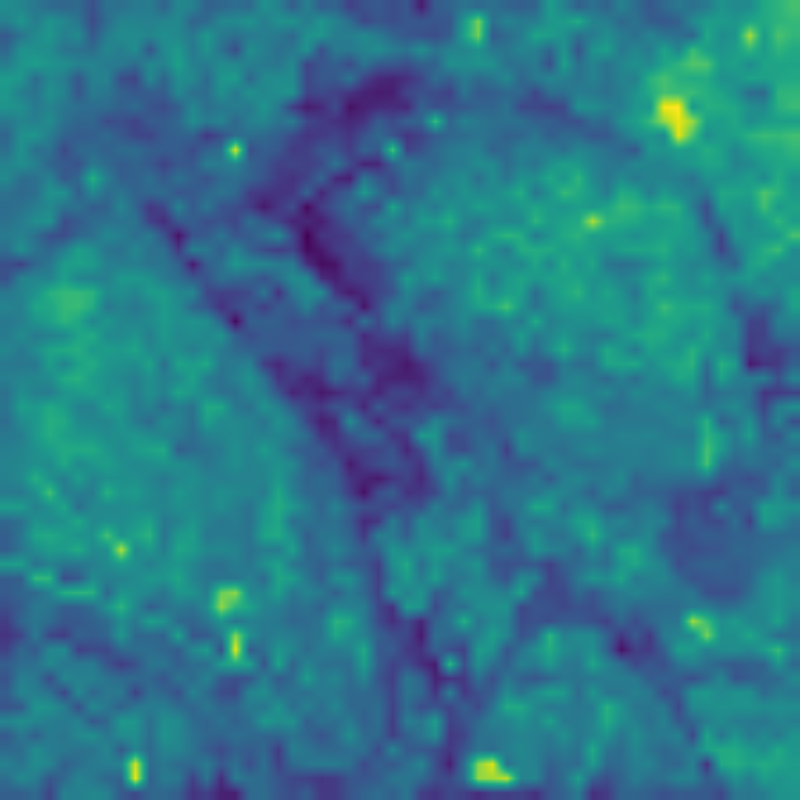}
        \caption{Feature Activation: ($56\times56$)}
    \end{subfigure}

    \caption{Spatial feature representations extracted by the shared, frozen DINOv2 backbone across both input modalities. (Top) Raw input imagery; (Bottom) corresponding patch-token activation magnitude heatmaps.}
    \label{fig:shared_backbone_features}
\end{figure}

\subsubsection{Classification Head}
For terrain classification, the backbone's $384$-dimensional CLS token, which aggregates global context across the $256$ patch tokens, is passed through a two-layer fully connected network with ReLU~\cite{relu_vnair_geoffrey} activation, followed by a softmax that produces the predicted probability vector over the seven terrain classes. The corresponding equations are provided in Supplementary Section~S2.

\begin{figure*}[t]
    \centering
    \begin{subfigure}[t]{0.49\textwidth}
        \includegraphics[width=\linewidth]{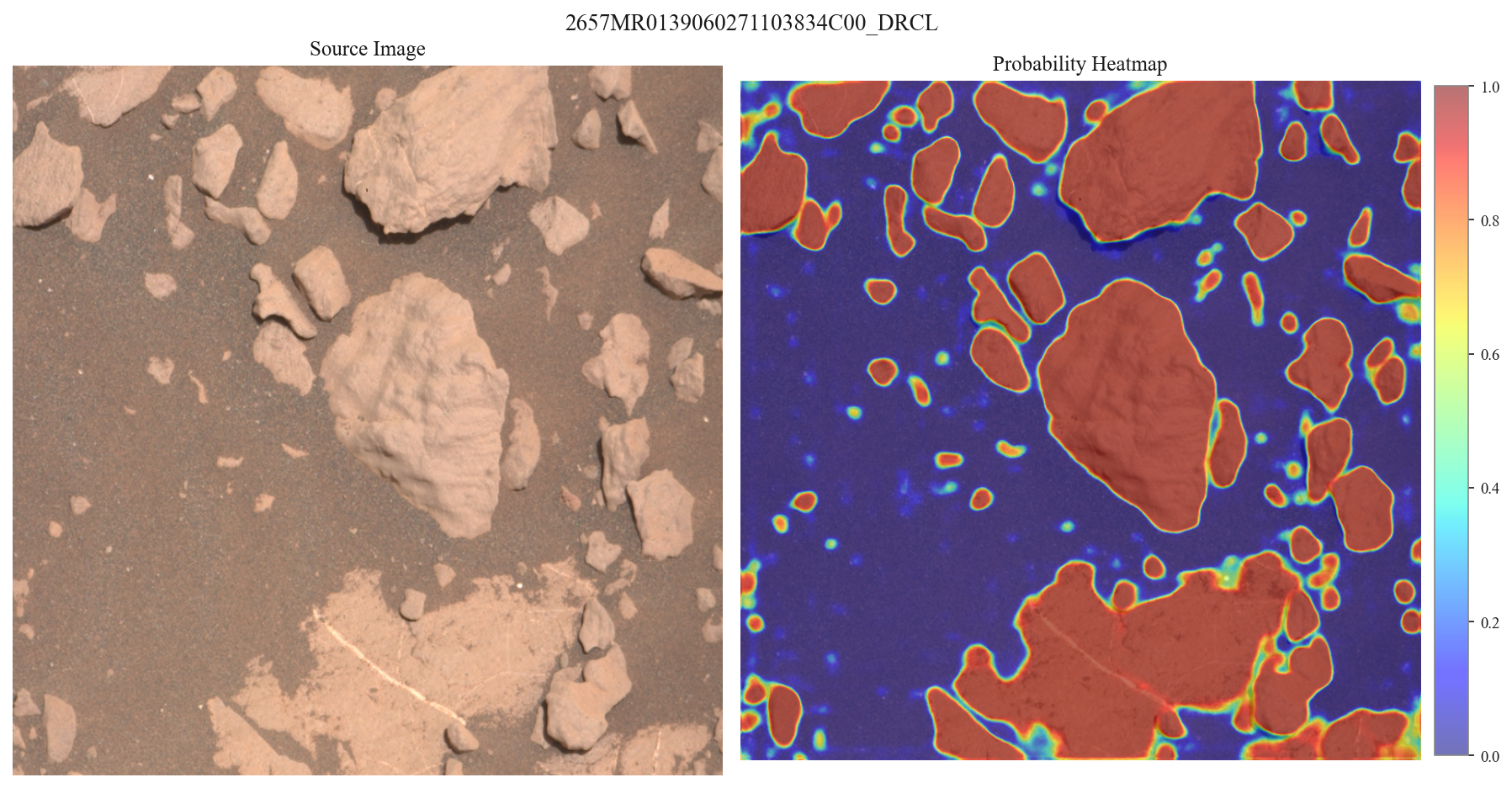}
        \caption{High IoU example (IoU $= 0.90$)}
    \end{subfigure}
    \hfill
    \begin{subfigure}[t]{0.49\textwidth}
        \includegraphics[width=\linewidth]{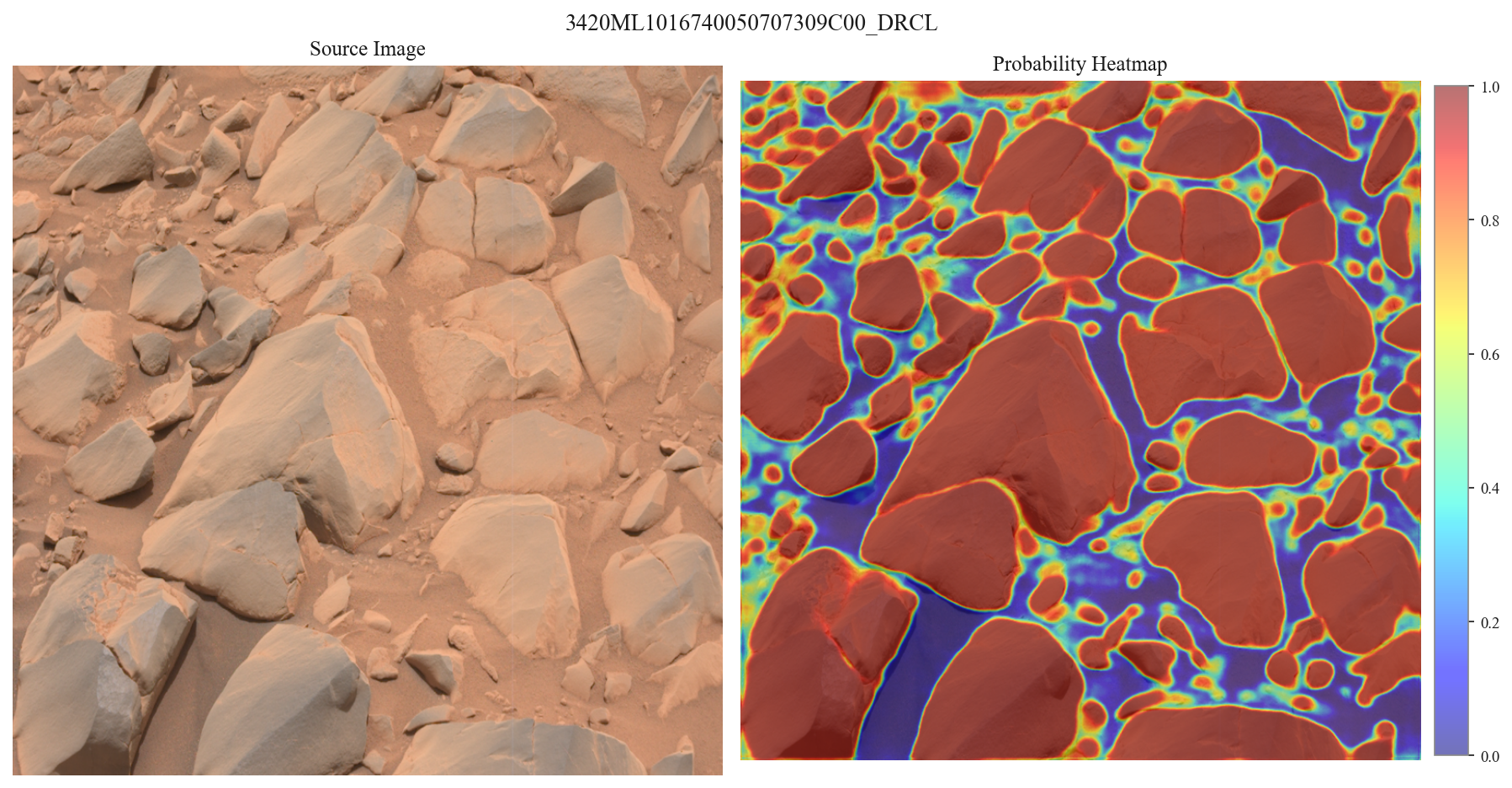}
        \caption{High-mid IoU example (IoU $= 0.80$)}
    \end{subfigure}
    \vspace{0.5em}
    \begin{subfigure}[t]{0.49\textwidth}
        \includegraphics[width=\linewidth]{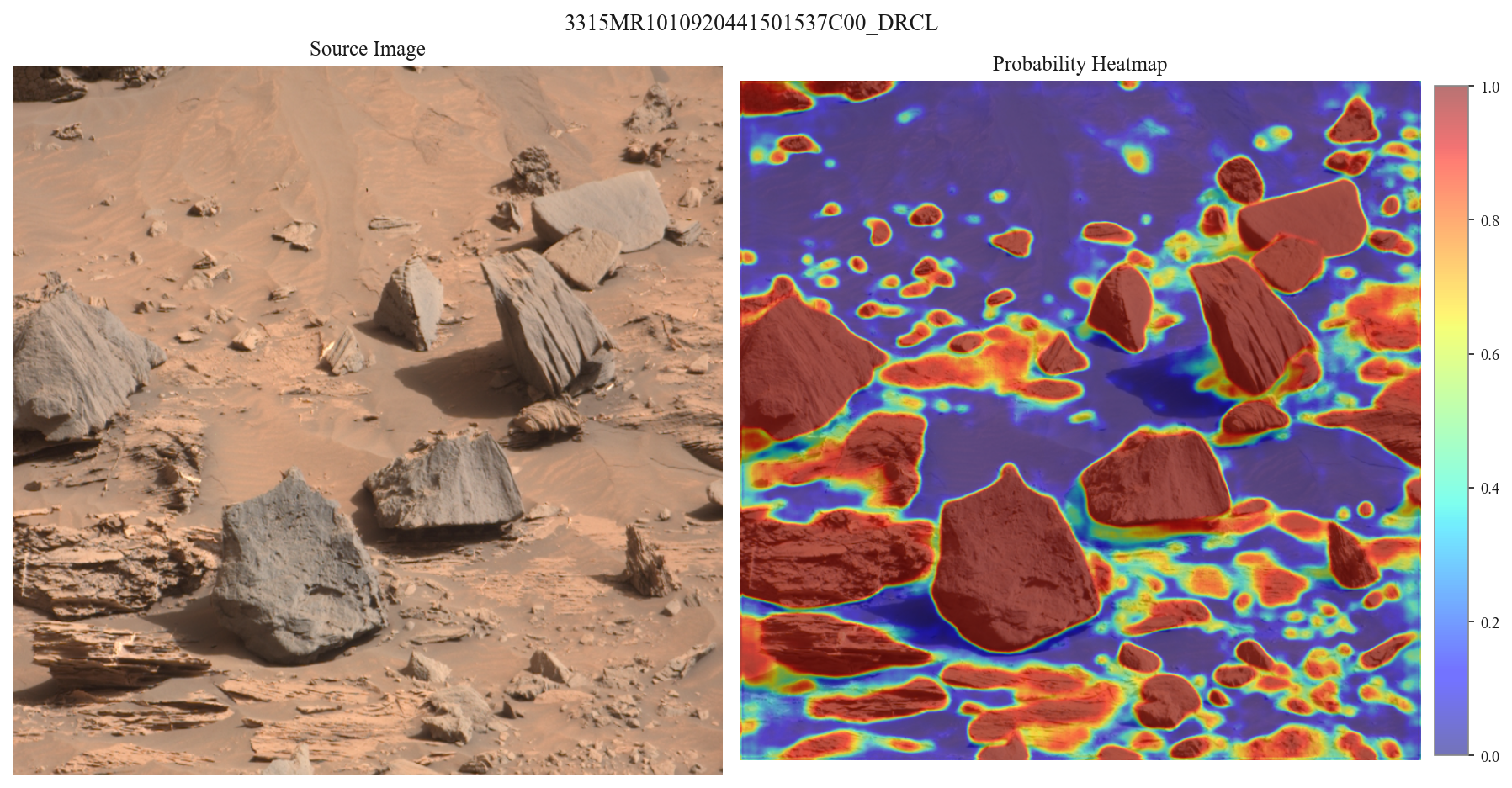}
        \caption{Mid-range IoU example (IoU $= 0.70$)}
    \end{subfigure}
    \hfill
    \begin{subfigure}[t]{0.49\textwidth}
        \includegraphics[width=\linewidth]{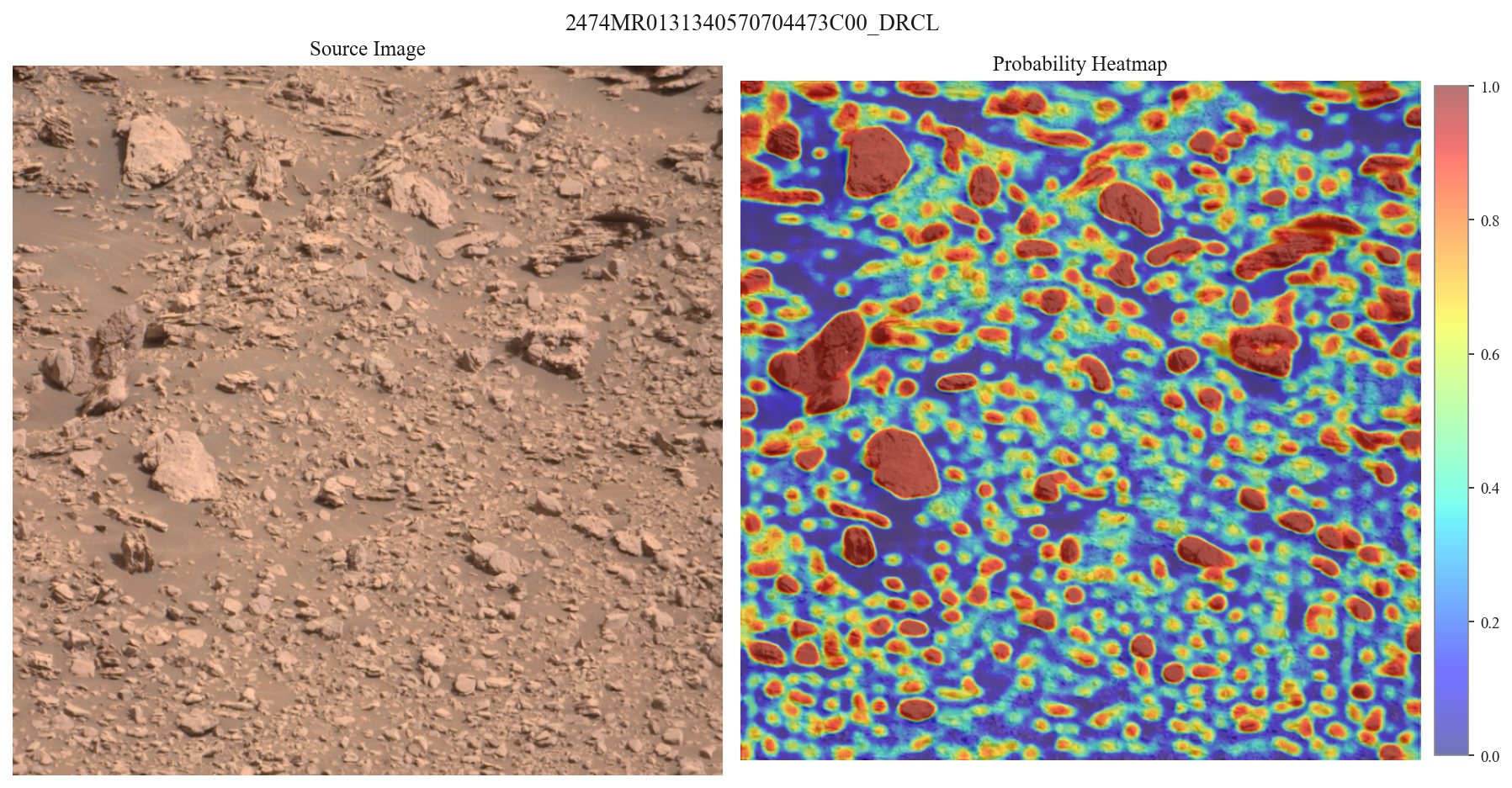}
        \caption{Low IoU example (IoU $= 0.50$), dominated by small rocks and pebbles}
    \end{subfigure}
    \caption{Qualitative segmentation results across the IoU distribution on the held-out cross-sol test set (Sol~$\geq$~2305). Each panel shows the source MSL Mastcam image (left) and the predicted boulder probability heatmap (right), with warmer colors indicating higher predicted boulder likelihood. (a), (b), and (c) illustrate strong segmentation of large boulders in both densely and sparsely distributed arrangements; (d) illustrates the dominant failure mode such as missed small rocks and pebbles, which has negligible impact on rover navigation.}
    \label{fig:qualitative_results}
\end{figure*}

\subsubsection{Segmentation Head}
The segmentation head receives the $768 \times 56 \times 56$ backbone feature grid and produces a full-resolution binary boulder mask through four stages. A dual attention module first recalibrates the features: a Squeeze-and-Excitation (SE)~\cite{hu2018squeeze} channel attention block globally pools the spatial dimensions to learn per-channel importance weights, and a Convolutional Block Attention Module (CBAM)~\cite{woo2018cbam} spatial attention block then suppresses spatially irrelevant activations. Two double-convolution encoder blocks progressively compress the channel depth ($768 \to 128 \to 64$) while preserving spatial resolution, with the second block using dilated convolutions ($d = 2$) to expand the effective receptive field without downsampling. Three successive upsample-and-refine decoder stages ($64 \to 32 \to 16 \to 16$) progressively recover spatial resolution, with auxiliary supervision heads attached at the first two decoder stages to produce the intermediate outputs used for deep supervision during training. Finally, a $1 \times 1$ convolution maps the decoder output to a single-channel logit map, which is bilinearly upsampled to a binary boulder mask at the original $784 \times 784$ MSL input resolution. The full layer-by-layer equations are provided in Supplementary Section~S2.

\subsubsection{Training Objectives}

The classification and segmentation heads are trained independently, each with its own loss function and optimizer.
 
The classification head is optimized using standard cross-entropy loss:
 
\begin{equation}
    \mathcal{L}_{\text{cls}} = -\sum_{i=1}^{7} y_i \log \hat{y}_i
\end{equation}
 
The segmentation head is optimized using a combined BCE, Dice, and boundary-aware loss on the main output, with annealed auxiliary supervision at the first two decoder stages:
 
\begin{align}
    \mathcal{L}_{\text{seg}} &= \underbrace{\mathcal{L}_{\text{BCE}}(\alpha)
        + \mathcal{L}_{\text{Dice}} + \mathcal{L}_{\text{Boundary}}}_{\text{main output}} \notag \\
        &\quad + \lambda_{\text{aux}}(t) \sum_{k=1}^{2}
          \left(\mathcal{L}^{(k)}_{\text{BCE}}(\alpha)
              + \mathcal{L}^{(k)}_{\text{Dice}}\right)
\end{align}
 
where $\alpha = 1.2$ is a positive-class weight, $\mathcal{L}_{\text{Boundary}}$ is a Laplacian-based boundary-weighted BCE term emphasizing mask edges, and $\lambda_{\text{aux}}(t)$ decays linearly from $0.3$ to $0$ over the first half of training. Full implementation details and hyperparameters are provided in Supplementary Section~S2.

\begin{figure*}
    \centering
    \includegraphics[width=0.8\textwidth]{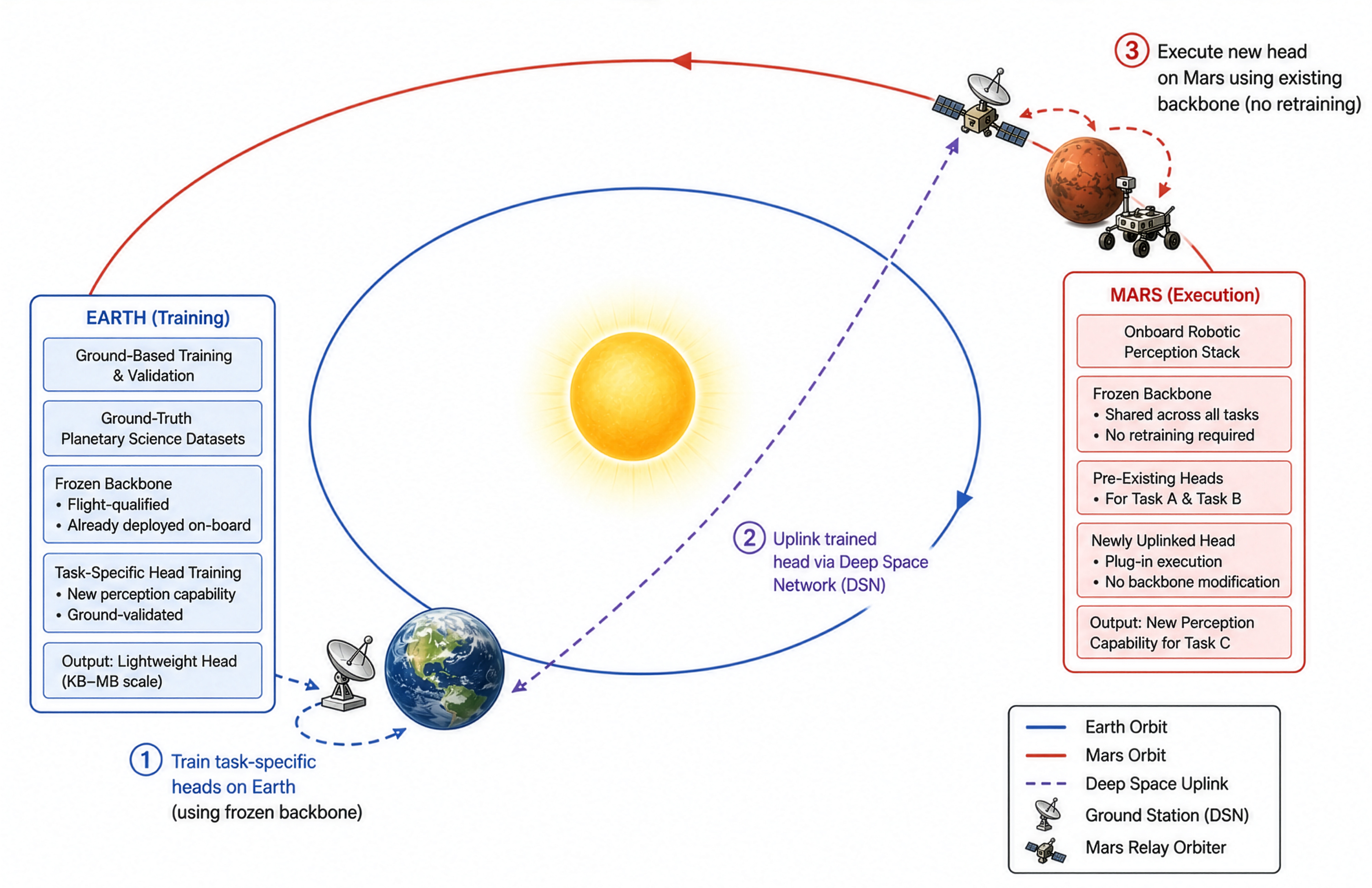}
    \caption{The Modular Uplink Principle. New perception capabilities are added to a deployed rover in three steps: (1)~a task-specific head is trained and validated on Earth against a copy of the frozen, flight-qualified backbone, yielding a lightweight (KB--MB scale) module; (2)~the trained head is uplinked via the Deep Space Network (DSN) and relayed to the surface by a Mars relay orbiter; and (3)~the head executes onboard as a plug-in to the existing frozen backbone alongside previously deployed heads, enabling a new capability without retraining or modifying the backbone or onboard flight software.}
    \label{fig:modular_uplink_principle}
\end{figure*}

\section{Experimental Evaluation and Results}\label{sec:results}

In this section, we present the experimental results and performance analysis of the proposed MANTLE model on both classification and segmentation tasks. We evaluate the classification head on a held-out test set using validation accuracy and a confusion matrix to characterize learning stability. For the segmentation head, we demonstrate boulder segmentation performance using qualitative and quantitative results on a held-out test set to demonstrate cross-sol generalization.

We also provide consolidated insights and observations from these experiments, highlighting key challenges, model behavior, and areas for future improvement.

\subsection{Terrain Classification Evaluation}
The classification head of MANTLE was trained to distinguish seven Martian terrain types: \textit{crater}, \textit{dark dune}, \textit{slope streak}, \textit{bright dune}, \textit{impact ejecta}, \textit{Swiss cheese}, and \textit{spider}. Training used a stratified 80/20 split across 2{,}450 labeled HiRISE patches, with a held-out test set of 336 unseen images (50 per class, except impact ejecta with 36 samples) selected to avoid spatial proximity to training tiles within the original full-resolution HiRISE files, guarding against train/test leakage. The model was trained for 100 epochs and achieved its best validation accuracy of \textbf{95.51\%} at epoch 97. On the 336-image held-out test set, the classifier achieved an accuracy of \textbf{92.56\%}. 
 
The model generalized strongly across most terrain types, with near-perfect accuracy on \textit{bright dune} and \textit{Swiss cheese}. The most frequent confusions occurred between \textit{impact ejecta}, \textit{spider}, and \textit{crater} are classes that share overlapping morphological signatures, such as radial patterns and shadowing effects, consistent with known ambiguities in Martian orbital geology. 
The normalized confusion matrix, provided in Supplementary Section~S3, reflects high accuracy and low inter-class confusion for most of the seven classes, though the per-class metrics in Supplementary Table~S3 show weaker performance on \textit{impact ejecta} (recall 0.72). 

Full per-class precision, recall, F1-score, and confusion matrix analysis are provided in Supplementary Section~S3.

\begin{table*}[t]
\centering
\caption{Measured parameter counts and serialized sizes: shared backbone vs.\ the two MANTLE task-specific heads. INT8 sizes are estimated at 1 byte per parameter.}
\label{tab:head_size_comparison}
\begin{tabular}{lcccc}
\toprule
Component & Task & Parameters & FP32 Size & INT8 Size (est.) \\
\midrule
Shared backbone (DINOv2 ViT-S/14) & Feature extraction & $2.2 \times 10^7$ & ${\sim}85$ MB & ${\sim}22$ MB \\
Classification head (MANTLE) & Terrain classification & $100{,}359$ & ${\sim}0.4$ MB & ${\sim}0.1$ MB \\
Segmentation head (MANTLE) & Boulder segmentation & $759{,}013$ & ${\sim}2.9$ MB & ${\sim}0.76$ MB \\
\bottomrule
\end{tabular}
\end{table*}

\subsection{Boulder Segmentation Evaluation}

The segmentation head of MANTLE was trained on the MSL Boulder Dataset using the frozen DINOv2 ViT-S/14 backbone, achieving a best validation IoU of \textbf{0.753} on the Sol~0017--2302 split. Training details and configuration hyperparameters are provided in Supplementary Sections~S2 and~S4. 

To evaluate cross-sol generalization, the model was tested on a held-out set of 53 images drawn exclusively from Sol~$\geq$~2305, comprising traverse segments with different terrain morphology, dust accumulation, and lighting conditions compared to the training range. The preprocessing pipeline was applied identically to the test set to ensure no distributional mismatch. The model achieved a test IoU of \textbf{0.735}, a modest decrease of $0.018$ from validation, demonstrating that the learned features generalized well to
previously unseen terrain. Of the 53 test images, $33$ ($62.3\%$) achieved an IoU of $0.70$ or higher, with the lowest-performing cases dominated by small rocks and pebbles, which pose negligible risk to rover traversal. The model's strongest performance was concentrated on the larger, traversal-relevant boulders that constitute genuine navigation hazards. The full per-image IoU distribution is provided in Supplementary Section~S4. 

To complement the quantitative results, \autoref{fig:qualitative_results} presents per-pixel prediction confidence heatmaps overlaid on representative test images from Sol~$\geq$~2305.

\section{Modular Uplink Principle}\label{sec:mup}

\subsection{Problem with Current Systems}
Modern planetary rovers rely on tightly integrated perception pipelines, where each computer-vision capability such as terrain classification, obstacle detection, novelty discovery, or science target identification is implemented as a separate model or subsystem. Updating or expanding these capabilities on active missions is challenging due to limited uplink bandwidth, short
communication windows, and the impracticality of transmitting large neural models. As a result, missions rarely expand their perception systems after landing, limiting long-term adaptability and slowing scientific and operational innovation.

\subsection{Proposed Approach}

The Modular Uplink Principle proposes a fundamental shift: only the shared backbone remains persistent onboard the rover, while new perception capabilities are delivered as lightweight, plug-in heads trained on Earth (\autoref{fig:modular_uplink_principle}). Under this principle, a single frozen backbone is uplinked once and reused for all future tasks; new capabilities such as classification, segmentation, novelty detection, and hazard mapping are implemented as compact task-specific heads and uplinked as small modules without modifying the backbone or disturbing flight-qualified onboard software. 
\autoref{tab:head_size_comparison} quantifies this separation with the measured MANTLE components: the task-specific heads introduce only $10^5$--$10^6$ parameters compared to $2.2 \times 10^7$ for the backbone, roughly 1.5--2.3 orders of magnitude fewer and translating to sub-megabyte to few-megabyte uplink payloads (${\sim}0.1$--$2.9$ MB), reducible further with quantization, making incremental capability updates practical under deep-space communication constraints.

\subsection{Use-Cases}

This principle enables several high-impact mission scenarios:
\begin{itemize}
    \item Post-landing capability expansion: New scientific priorities such as mineral phase detection, dust devil identification, or frost mapping can be supported by uplinking specialized heads without updating the backbone.
    \item Hazard detection and adaptive science targeting: Newly identified hazards or science targets such as fresh boulder fields, outcrop boundaries, hydrated mineral signatures, or lithologic contacts can be addressed via newly trained segmentation or detection heads uplinked mid-mission.
    \item Sample-caching and follow-on missions: Heads can be uplinked to detect cached sample tubes, rendezvous markers, or engineered artifacts necessary for Mars Sample Return.
\end{itemize}

\section{Conclusion}\label{sec:conclusion}
In this work, we presented MANTLE, a multi-task perception framework for in-situ Martian exploration built around a shared, frozen DINOv2 backbone with lightweight task-specific heads, trained on two newly curated high-resolution datasets: a balanced seven-class HiRISE landform dataset and a pixel-accurate MSL boulder segmentation dataset spanning 840 sols. The classification head achieved a test accuracy of 92.56\%, and the segmentation head reached a validation IoU of 0.753 with only a modest drop to 0.735 on a temporally held-out test set, demonstrating strong cross-sol generalization. Formalized through the Modular Uplink Principle, these results show that scientific terrain understanding and operational hazard assessment can coexist within a single extensible architecture in which new capabilities are trained on Earth and delivered as KB--MB-scale head uplinks without retraining or modifying the onboard model.

The most important next step is to validate the uplink principle end-to-end: quantizing the trained heads to 8-bit precision, measuring the resulting accuracy and IoU against the FP32 baselines reported here, and benchmarking inference on flight-like hardware such as Snapdragon-class processors of the kind flown on Ingenuity. Beyond that, future work will extend the framework with additional heads, such as novelty detection, science target identification, and cached-sample localization, each trained against the same frozen backbone and validated under realistic uplink budgets, moving toward surface explorers that continue to acquire new perception skills over a mission's lifetime.

\section*{Acknowledgments}

We wish to thank Steven Lu for providing traces to locate landforms within the HiRISE Landform Dataset in full-resolution images. Part of this research was carried out at the Jet Propulsion Laboratory, California Institute of Technology, under a contract with the National Aeronautics and Space Administration (80NM0018D0004).

This work has been submitted to a journal for possible publication. Copyright may be transferred without notice, after which this version may no longer be accessible.

\section*{Disclosures}
The authors declare no conflicts of interest.

\bibliographystyle{IEEEtran}
\bibliography{reference}

\end{document}